\documentclass[journal]{IEEEtran}

\usepackage{graphicx}  

\usepackage{amsmath}   

\usepackage{epsfig}
\usepackage{algorithm}
\usepackage{algpseudocode}
\usepackage{url}

\newtheorem{theorem}{\indent Theorem}

\newtheorem{definition}{\indent Definition}

\newtheorem{remark}{Remark}

\begin{document}
%
\title{Fidelity-based Probabilistic Q-learning for Control of Quantum Systems}

\author{Chunlin~Chen,
        Daoyi~Dong,
        Han-Xiong~Li,
        Jian Chu,
        Tzyh-Jong~Tarn
\thanks{This work was supported by the Natural Science Foundation of China (Nos.61273327 and 61374092), Australian Research Council's Discovery Projects funding scheme (project DP130101658) and the Fundamental Research Funds for the Central Universities.}
\thanks{C. Chen is with the Department of Control and System Engineering, Nanjing
University, Nanjing 210093, China and with the Department of
Chemistry, Princeton University, Princeton, NJ 08544, USA}
\thanks{D. Dong is with the School of
Information Technology and Electrical Engineering, University of New
South Wales at the Australian Defence Force Academy,
Canberra, ACT
2600, Australia
and
with the Institute of Cyber-Systems
and Control,
Zhejiang University, Hangzhou 310027, China}
\thanks{J. Chu is
with the Institute of Cyber-Systems
and Control, State Key Laboratory of Industrial Control Technology,
Zhejiang University, Hangzhou 310027, China.}
\thanks{H.X. Li is with the Department of
Systems Engineering and Engineering Management, City University of Hong Kong, Hong Kong,
China.}
\thanks{T.J. Tarn is with the Department of Electrical and
Systems Engineering, Washington University in St. Louis, St. Louis,
MO 63130 USA.}}


\maketitle

\begin{abstract}
The balance between exploration and exploitation is a key problem
for reinforcement learning methods, especially for Q-learning. In
this paper, a fidelity-based probabilistic Q-learning (FPQL)
approach is presented to naturally solve this problem and applied
for learning control of quantum systems. In this approach, fidelity is
adopted to help direct the learning process and the probability
of each action to be selected at a certain state is updated
iteratively along with the learning process, which leads to a
natural exploration strategy instead of a pointed one with
configured parameters. A probabilistic Q-learning (PQL) algorithm is
first presented to demonstrate the basic idea of probabilistic
action selection. Then the FPQL algorithm is presented for learning
control of quantum systems. Two examples (a spin-$\frac{1}{2}$
system and a $\Lambda$-type atomic system) are demonstrated to test
the performance of the FPQL algorithm. The results show that FPQL
algorithms attain a better balance between exploration and
exploitation, and can also avoid local optimal policies and
accelerate the learning process.
\end{abstract}

\begin{keywords}
Fidelity, probabilistic Q-learning, quantum control, reinforcement
learning.
\end{keywords}

\section{Introduction}
\PARstart{R}{einforcement} learning (RL) \cite{Sutton and Barto
1998} is an important approach to machine learning, control
engineering, operations research, etc. RL theory addresses the
problem of how an active agent can learn to approximate an optimal
behavioral strategy while interacting with its environment. RL
algorithms, such as the temporal difference (TD) algorithms
\cite{Sutton 1988} and Q-learning algorithms \cite{Watkins and Dayan
1992}, have been deeply studied in various aspects and widely used
in intelligent control and industrial applications \cite{Kondo and
Ito 2004}-\cite{Yang and Jagannathan 2012}. However, there exist
several difficulties in developing practical applications from RL
methods. These difficult issues include tradeoff between exploration
and exploitation, function approximation methods and speed-up of the
learning process. Hence, new ideas are necessary to improve
reinforcement learning performance. In \cite{Dong et al 2008a}, we
considered two features (i.e., quantum parallelism and probabilistic
phenomena) from the superposition of probability amplitudes in
quantum computation \cite{Nielsen and Chuang 2000} to improve TD
learning algorithms \cite{Dong et al 2008a}, \cite{Chen et al 2006},
\cite{Dong et al 2008b}. Inspired by \cite{Dong et al 2008a}, this
paper focuses on only the probabilistic essence of decision-making
in Q-learning \cite{Chen et al 2008b} with fidelity-directed
exploration strategy, and propose a fidelity-based probabilistic
Q-learning method for the control design of quantum systems.

We focus on exploration strategies (i.e., action selection methods),
which contribute to better balancing between exploration and
exploitation and have attracted more and more attention from
different areas \cite{Guo et al 2004}-\cite{Kim et al 2012}. For
Q-learning, exploitation (i.e., the greedy action selection) occurs
if the action selection strategy is based on only current values of
the state-action pairs. In most optimization problems, this will
lead to locally optimal policies, possibly differing from a globally
optimal one. In contrast, exploration is a strategy based on the
assumption that the agent selects a non-optimal action in the
current situation and obtains more knowledge about the problem. This
knowledge allows the agent to neglect the locally optimal policies
and to reach the globally optimal one. However, excessive
exploration will drastically decrease the performance of a learning
algorithm. Generally in a reinforcement learning process without
prior knowledge or training data, most of existing exploration
strategies are undirected exploration. Up to now, there have existed
two main types of undirected exploration strategies:
$\epsilon$-greedy strategy and randomized strategy \cite{Sutton and
Barto 1998}, where the randomized strategy includes such methods as
Boltzmann exploration ( i.e., Softmax method) and simulated
annealing (SA) method \cite{Sutton and Barto 1998}, \cite{Guo et al
2004}. These exploration strategies usually suffer from difficulties
in balancing between exploration and exploitation, and providing an
easy mechanism of parameter setting. Hence, the aim of this paper is
to propose a novel fidelity-based probabilistic action selection
method to improve Q-learning algorithms.

In this approach, we systematically investigate the use of
probabilistic action selection mechanism (e.g., see \cite{Bowling
and Veloso 2001} and Section 6.6 in \cite{Sutton and Barto 1998})
to dynamically balance the exploration and exploitation in
reinforcement learning. Furthermore, a fidelity-based
probabilistic Q-learning algorithm is presented for learning
control of quantum systems. The development of control design
approaches for quantum systems is a key task for powerful quantum
information technology \cite{Nielsen and Chuang 2000}, \cite{Dong
and Petersen 2010IET}-\cite{Chen et al 2012}. Unique
characteristics of quantum systems (e.g., ultrafast dynamics,
measurement destroying quantum states) make open loop strategies
competitive \cite{Dong and Petersen 2010IET}, \cite{Wiseman and
Milburn 2009}. Here we employ a reinforcement learning approach to
design control laws for a class of quantum control problems where
the set of control fields is given. Once the control sequence is
obtained by learning, the corresponding control fields can be
applied to the quantum system to be controlled. The method is very
useful for quantum systems since it is an important objective to
find control laws for complex quantum control problems when we
have limited resources. However, if we employ a basic
reinforcement learning algorithm, the direction of achieving the
objective is always delayed due to the lack of feedback
information during the learning process unless the agent reaches
the target state. Hence, the learning process is time-consuming
and the agent learns very slowly, which impedes the applications
of reinforcement learning methods to complex learning problems
with large learning space. In quantum information theory, the
``closeness" between two quantum states can be measured by
fidelity \cite{Nielsen and Chuang 2000}, \cite{Bason et al 2011},
\cite{Cozzini et al 2007}. The more similar two quantum states
are, the greater the fidelity between the two states is. For
example, the fidelity of two identical quantum states is usually
defined as 1 and the fidelity of two orthogonal quantum states is
defined as 0. The fidelity $F$ between two quantum states
corresponds to a non-negative number $F\in [0, 1]$. Hence, the
information of fidelity can be sent back to the learning system
and help speed up the learning process as a global direction
signal to avoid getting lost. Recent research on quantum control
landscapes provides a theoretical footing for the development of
new learning algorithms using the information of fidelity
\cite{Rabitz et al 2004}, \cite{Chakrabarti and Rabitz 2007}.
Numerical examples show that the fidelity-based probabilistic
Q-learning method has improved performance for learning control of
quantum systems.

This paper is organized as follows. Section II
introduces the basic Q-learning method and the existing exploration
strategies. In Section III, the
probabilistic action selection strategy is presented. Then a
probabilistic Q-learning (PQL) algorithm and a fidelity-based PQL
(FPQL) algorithm are proposed and analyzed aiming at speeding up the
learning process. In Section IV, the FPQL algorithm is applied
to learning control of two typical classes of quantum
systems (a spin-$\frac{1}{2}$ system and a $\Lambda$-type atomic
system), respectively. Conclusions are given in Section V.

\section{Q-learning and Exploration Strategy}
Q-learning can acquire optimal control policies from delayed
rewards, even when the agent has no prior knowledge of the
environment. For the discrete case, a Q-learning algorithm assumes
that the state set $S$ and action set $A$ can be divided into
discrete values. At a certain step $t$, the \emph{agent} observes
the state $s_t$, and then chooses an action $a_t$. After executing
the action, the agent receives a reward $r_{t+1}$, which reflects
how good that action is (in a short-term sense). The state will
change into the next state $s_{t+1}$ under action $a_t$. Then the
agent will choose the next action $a_{t+1}$ according to the best
known knowledge. The goal of Q-learning is to learn a policy
$\pi:S\times \cup_{i\in S}A_{(i)}\to [0,1]$, so that the expected
sum of discounted rewards for each state will be maximized:
\begin{eqnarray}
Q_{(s,a)}^{\pi}
=r_{s}^{a}+\gamma\sum_{s'}p_{ss'}^{a}\sum_{a'}p^{\pi}(s',a')Q_{(s',a')}^{\pi}
\end{eqnarray}
where $\gamma \in [0,1)$ is a discount factor,
$p_{ss'}^{a}=Pr\{s_{t+1}=s'|s_t=s,a_t=a\}$ is the probability for
state transition from $s$ to $s'$ with action $a$,
$p^{\pi}(s',a')$ is the probability of selecting action $a'$ for
state $s'$ under policy $\pi$ and
$r_{s}^{a}=E\{r_{t+1}|s_t=s,a_t=a\}$ is an expected one-step
reward. $Q_{(s,a)}$ is called the value function of state-action
pair $(s,a)$. Let $\alpha_{t}$ be the learning rate. The one-step
updating rule of Q-learning may be described as:
\begin{equation}\small
Q(s_t,a_t)\leftarrow(1-\alpha_{t})Q(s_t,a_t)+\alpha_{t}(r_{t+1}+\gamma
\max_{a'}Q(s_{t+1},a')).
\end{equation}
The optimal value function $Q_{(s,a)}^{*}$ satisfies the Bellman equation
\cite{Sutton and Barto 1998}:
\begin{eqnarray} \label{Bellman}
Q_{(s,a)}^{*}=\max_{\pi}Q_{(s,a)}=r_{s}^{a}+\gamma\sum_{s'}p_{ss'}^{a}\max_{a'}Q_{(s',a')}^{\pi}.
\end{eqnarray}
More details about Q-learning can be found in \cite{Sutton and Barto
1998}, \cite{Watkins and Dayan 1992}.

To efficiently approach the optimal policy
\[\pi^{*}=\arg
\max_{\pi}Q_{(s,a)}^{\pi} (\forall s\in S),\] where $\pi^{*}$ is the
optimal policy when $Q_{(s,a)}^{\pi}$ is maximized, Q-learning
always needs a certain exploration strategy (i.e., the action
selection method). One widely used action selection method is
$\epsilon$-greedy $(\epsilon\in [0,1))$ \cite{Sutton and Barto
1998}, where the optimal action is selected with probability $(1 -
\epsilon)$ and a random action is selected with probability
$\epsilon$. Sutton and Barto \cite{Sutton and Barto 1998} have
compared the performance of RL algorithms with different $\epsilon$
and have shown that a nonzero $\epsilon$ is usually better than
$\epsilon=0$ (i.e., the blind greedy strategy). In addition, the
exploration probability $\epsilon$ can be reduced over time, which
moves the learning from exploration to exploitation. The
$\epsilon$-greedy method is simple and effective, but it has the
drawback that when the learning system explores it chooses equally
among all actions. This means that the learning system makes no
difference between the worst action and the next-to-best action.
Another problem is that it is difficult to choose a proper parameter
$\epsilon$ for the optimal balancing between exploration and
exploitation.

Another kind of action selection methods is randomized strategies,
such as the Softmax method \cite{Sutton and Barto 1998} and the
simulated annealing method \cite{Guo et al 2004}. Such methods use a
positive parameter $\tau$ called a temperature and choose an action
$a$ with the probability proportional to $e^{Q(s,a)/\tau}$. Compared
with the $\epsilon$-greedy method, the ``best" action is still given
the highest selection probability, but all the others are ranked and
weighted according to their estimated $Q$-values. It can also move
from exploration to exploitation by adjusting the ``temperature"
parameter $\tau$. It is natural to sample actions according to this
distribution, but it is very difficult to set and adjust a good
parameter $\tau$ and may converge slowly. Another shortcoming is
that it does not work well when the $Q$-values of the actions are
close and the best action cannot be separated from the others.
Moreover, when the parameter $\tau$ is reduced over time to acquire
more exploitation, there is no effective mechanism to guarantee
re-exploration when necessary.

\begin{figure}
\centering
\includegraphics[width=3.3in]{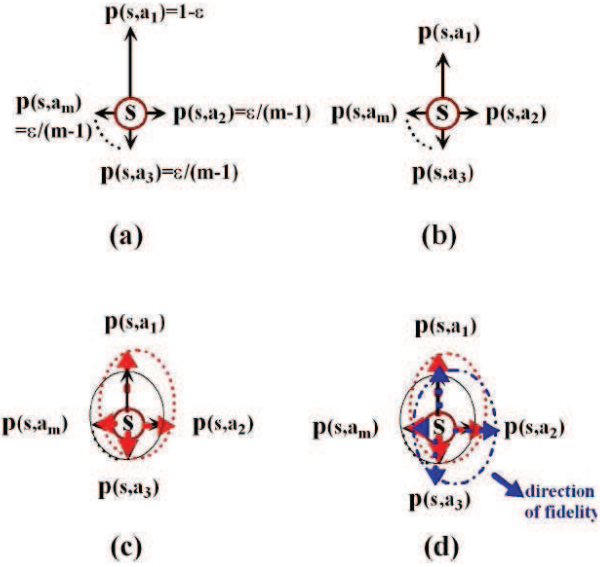}
\caption{Illustration of the idea of probabilistic action selection
method and the effect of fidelity. (a) $\epsilon$-greedy method; (b)
Softmax method; (c) basic probabilistic action selection method; (d)
fidelity-based probabilistic action selection method.}
\label{Expolicy}
\end{figure}

To sum up, these existing exploration strategies usually suffer
from difficulties in balancing between exploration and
exploitation, setting appropriate parameters and providing an
effective mechanism of re-exploration. Here, we present a novel
fidelity-based probabilistic Q-learning algorithm where a
probabilistic action selection method is used as more effective
exploration strategies to improve the performance of Q-learning
for complex learning control problems. Compared with
$\epsilon$-greedy and Softmax methods, Fig. 1 shows the
illustration of the ideas of straightforward probabilistic action
selection method and the fidelity-based one, where a longer line with an arrow indicates
a higher probability. As shown in Fig.
1(a), $\epsilon$-greedy method uses a prefixed exploration policy
and the action $a_{1}$ with the maximum of Q-value ($Q(s,a_{1})$)
is selected with the probability of $(1-\epsilon)$ and all the
other actions ($a_{2}\sim a_{m}$) are select with the probability
of $\epsilon/(m-1)$, respectively. Using Softmax method (Fig.
1(b)), the action $a_{i}, i=1,2,\ldots, m$, is selected with the
probability of
$\frac{e^{Q(s,a_{i})/\tau}}{\sum_{j=1}^{m}e^{Q(s,a_{j})/\tau}}$.
Fig. 1(c) shows that the action selection probability distribution
is dynamically updated (denoted with the dashed line) along with
the learning process instead of being computed from the estimated
Q-values and a temperature parameter. In Fig. 1(d), the fidelity is
used to direct the probability distribution and to strengthen the
learning effects with a regulation on the updating process. These
different exploration policies will be further explained and
compared from a point of view of physical mechanism (as shown in
Fig. 4) after the PQL and FPQL method are systematically presented
in the next section.

\section{Fidelity-based Probabilistic Q-learning}

\subsection{Probabilistic Action Selection and Reinforcement Strategy}

Inspired by the work in \cite{Dong et al 2008a}, \cite{Chen et al
2006}, we reformulate the action selection strategy in a unified
probabilistic representation where the action selection
probability distribution is updated based on the reinforcement
strategy. The discrete probability
distribution on the state-action space is defined as follows.

\begin{definition}
The \emph{probability distribution on the state-action space}
(discrete case) of a RL problem is characterized by a probability
mass function defined on the state set $S$ and the action set
$A=\bigcup_{s\in S}A_{(s)}$, where $A_{(s)}$ is the set of all the
permitted actions for state $s$. For any $s\in S$ and $a \in
A_{(s)}$, the probability mass function is defined as
$p(s,a)\geq0$ and for a certain state $s$, it satisfies
\begin{equation}
\sum_{a \in A_{(s)}}p(s,a)=1 .
\end{equation}
\end{definition}

Suppose the state-action space is $ S\times A $, where
\begin{equation}
S=\{s_{1},s_{2},\ldots ,s_{n}\}
\end{equation}
\begin{equation}
A=\bigcup_{s\in S}A_{(s)}=\{a_{1},a_{2}, \ldots,a_{m}\} .
\end{equation}
From Definition 1, the policy to be learned $\pi:S\times A \to
[0,1]$ can be represented using the probability distribution of the
state-action space
\begin{equation}
\pi: P^{\pi}=(p^{\pi}(s,a))_{n\times m}
\end{equation}
where $s \in S, a \in A$ and for a certain state $s$, the
probability distribution on the action set $A$ is
$p_{s}^{\pi}=\{p^{\pi}(s,a_{i})\}, i=1,2, \ldots, m$. The
look-up table for the Q-values and the probability distribution are
of the form
\begin{equation}
\left(
\begin{array}{ccccc}
 & \mathbf{a_1} & \mathbf{a_2} & \cdots & \mathbf{a_m} \\
\mathbf{s_1} & \left[ \begin{split}Q_{(s_1,a_1)}\\p^{\pi}_{(s_1,a_1)}\end{split}\right] & \left[ \begin{split}Q_{(s_1,a_2)}\\p^{\pi}_{(s_1,a_2)}\end{split}\right] & \cdots & \left[ \begin{split}Q_{(s_1,a_m)}\\p^{\pi}_{(s_1,a_m)}\end{split}\right]\\
\mathbf{s_2} & \left[ \begin{split}Q_{(s_2,a_1)}\\p^{\pi}_{(s_2,a_1)}\end{split}\right] & \ddots &  & \vdots \\
\vdots  & \vdots  &  & \ddots& \vdots\\
\mathbf{s_n} & \left[ \begin{split}Q_{(s_n,a_1)}\\p^{\pi}_{(s_n,a_1)}\end{split}\right] & \cdots & \cdots & \left[ \begin{split}Q_{(s_n,a_m)}\\p^{\pi}_{(s_n,a_m)}\end{split}\right]\\
\end{array}\right).
\end{equation}

\begin{figure}
\centering
\includegraphics[width=3.3in]{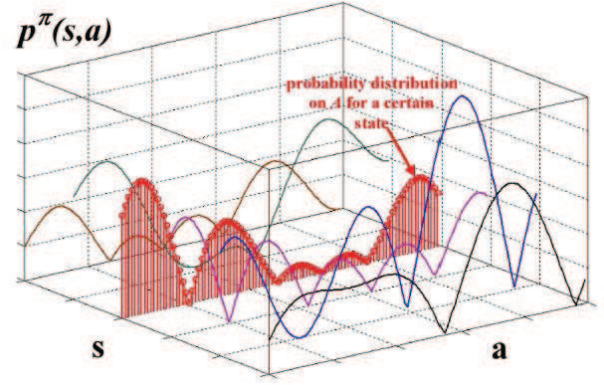}
\caption{3-D illustration of probabilistic distribution on the
state-action space for action selection under current policy
$\pi$.} \label{pdf}
\end{figure}

Fig. 2 shows a 3-D illustration of the probabilistic distribution
on the state-action space for action selection under current
policy $\pi$. In the probabilistic action selection method, one selects an action $a$
(under policy $\pi$) at a certain state $s$ with the probability
according to the probability distribution on the action set $A$,
i.e.,
\begin{equation}
a_{s}^{\pi} = f^{\pi}(s) =
\begin{cases}
  a_{1}   &  \text{with probability} \ p^{\pi}(s,a_{1}) \\
  a_{2}    &  \text{with probability} \ p^{\pi}(s,a_{2})    \\
  \vdots    &                  \\
  a_{m}      & \text{with probability} \ p^{\pi}(s,a_{m})   \\
\end{cases}
\end{equation}
Such a probabilistic action
selection method leads to a natural
probabilistic exploration strategy for Q-learning.

\begin{figure*}
\centering
\includegraphics[width=5.2in]{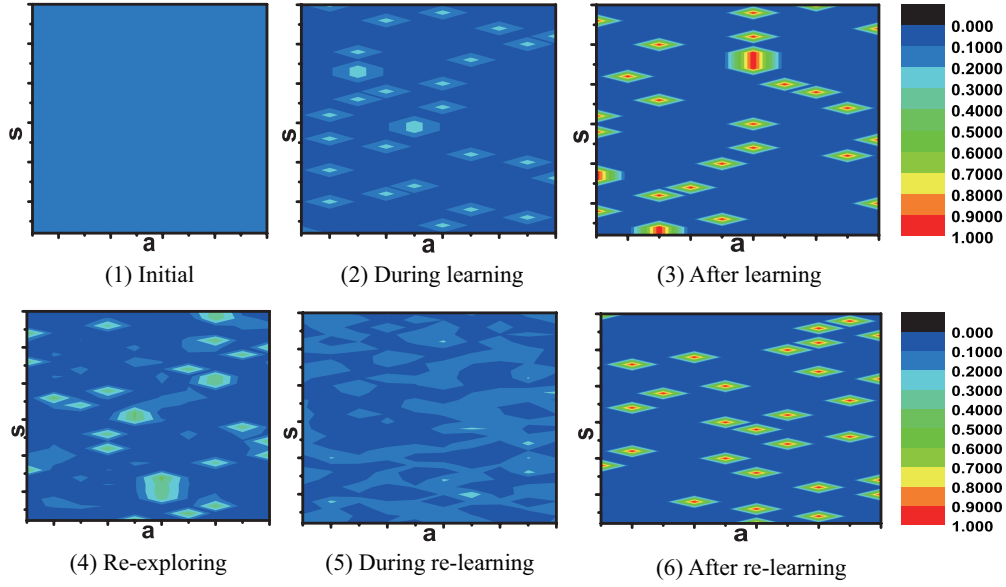}
\caption{Illustration of the evolution of the probabilistic
distribution for action selection
 on the state-action space during the
learning and re-exploring process.} \label{probability}
\end{figure*}

The goal of PQL is to learn a mapping from states to actions. The
agent needs to learn a policy $\pi$ to maximize the expected sum
of discounted reward for each state:
\begin{eqnarray}
Q_{(s, a)}^{\pi}
=\sum_{a\in A_{(s)}}p^{\pi}(s,a)[r_{s}^{a}+\gamma
\sum_{s'}p_{ss'}^{a}Q_{(s', a')}^{\pi}].
\end{eqnarray}
The one-step updating rule of PQL for $Q_{(s, a)}$ is the same as
that of QL
\begin{equation}\small \label{updaterule}
Q(s_t,a_t)\leftarrow(1-\alpha_{t})Q(s_t,a_t)+\alpha_{t}(r_{t+1}+\gamma
\max_{a'}Q(s_{t+1},a')).
\end{equation}
Besides the updating of $Q_{(s, a)}$, the probability distribution
is also updated for each learning step. After the execution of
action $a_{t}$ for state $s=s_{t}$, the corresponding probability
$p(s_{t},a_{t})$ is updated according to the immediate reward
$r_{t+1}$ and the estimated value of $Q_{(s', a')}$ for next state
$s'=s_{t+1}$
\begin{equation}
p(s_{t},a_{t}) \leftarrow p(s_{t},a_{t})  +
k(r_{t+1}+\max_{a'}Q(s_{t+1},a')),
\end{equation}
where $k$ ($k\geq0$) is an updating step size and the probability
distribution of actions at state $s=s_{t}$ \{${p(s,a_{1}),
p(s,a_{2}), ..., p(s,a_{m})}$\} is normalized after each updating.
The parameter setting of $k$ is accomplished by experience and
generally can be set as the same as the learning rate
$\alpha_{t}$. The variation of $k$ in a relatively large range
will only slightly affect the learning process because the probability
distribution \{${p(s,a_{1}), p(s,a_{2}), ..., p(s,a_{m})}$\} is
normalized after each updating step. In particular, $k$ is set as
$0.01$ for all the experiments in this paper.

The evolution of the probability distribution of action selection on
the state-action space during the whole learning process is shown as
in Fig. 3, where the values of the action selection probabilities
are represented with different colors. The probability distribution
usually starts with an initial uniform one before learning (as shown
in Fig. 3(1)), i.e., for each state $s$ the probability distribution
of action selection is initialized as $\{p(s,a_{i})=\frac{1}{m},
i=1,2,\ldots,m\}$. It evolves with the learning and probability
updating process (a sample of the probability distribution during
the learning process is shown in Fig. 3(2)) and reaches an optimal
one when the learning process ends (Fig. 3(3)). Then the learned
policy is applied to the agent (or a control system) and at the same
time the learning system may still keep on-line learning
capability. If the environment changes the probability distribution
will be updated accordingly (Fig. 3(4)) and naturally trigger a
re-learning process for the learning system. Fig. 3(5) and Fig. 3(6)
show the process of re-learning for a new environment. The
characteristics of the proposed probabilistic exploration strategy
(as shown in Fig. 3) is very different from the traditional one,
e.g., the $\epsilon$-greedy method for Q-learning, where action
probability distribution for a certain state keeps constant as
shown in Fig. 1(a).

\begin{remark}
The tradeoff between exploration and exploitation is a specific
challenge of RL. Compared with existing exploration strategies,
such as $\epsilon$-greedy, Softmax and simulated annealing
methods, the probabilistic exploration strategy has the following
merits. (i) The learning algorithm possesses more reasonable
credit assignment using a probabilistic method and the action
selection method is more natural without too much difficulty for
parameter setting. The only parameter to be set is the step size
$k$. The parameter $k$ will not substantially affect the algorithm
performance, because the action selection probabilities for a
certain state are relative and will be normalized after each
updating step. (ii) The method provides a natural re-exploring
mechanism (as shown in Fig. 3), i.e., when the environment
changes, the policy also changes along with the on-line learning
process. Such a re-exploring mechanism is difficult to implement
for the existing exploration strategies (e.g., $\epsilon$-greedy).
For example, the value of $\epsilon$ is usually decreased along
with the learning process to exploit more after a lot of trials.
When the environment changes the value of $\epsilon$ should be
reset to avoid too much exploitation. However, it is difficult to
do so intelligently. Our scheme provides a straightforward and
natural approach for re-exploring mechanism. Although it may need
a little bit more physical memories for probability distribution
updating, it will not substantially degrade the algorithm, while
the performance improvement is more prominent.
\end{remark}

\subsection{Probabilistic Q-learning Algorithms}

The procedural form of a probabilistic Q-learning algorithm is
presented as \emph{Algorithm 1}. In this PQL algorithm, after
initializing the state and action we can choose $a_{t}$ according to
the action probability distribution at state $s_t=s$. Execute this
action and the system can give the next state $s_{t+1}=s'$,
immediate reward $r_{t+1}$ and the estimated next state-action
function value $Q(s_{t+1},a')$. $Q(s_t,a_t)$ is updated by the
one-step Q-learning rule. The updating of $p(s_t,a_{t})$ (the
probability of choosing $a_{t}$ at state $s_t$) is also carried out
based on $r_{t+1}$ and $Q(s_{t+1},a')$. Hence, in the PQL algorithm,
the exploration policy is accomplished through a probability
distribution over the action set for each state. When the agent
chooses an action at a certain state $s$, the action $a_{i}$ will be
selected with probability $p(s,a_{i})$ which is also updated along
with the value function updating.

\begin{algorithm}
\caption{Probabilistic Q-learning} \label{PQL}

\begin{algorithmic}[1]

\State Initialize $Q(s,a)$ arbitrarily

\State Initialize the policy \small{$\pi:
P^{\pi}=(p^{\pi}(s,a))_{n\times m}$} to be evaluated

\Repeat {\ (for each episode)}

\State Initialize $t=1$, $s_{t}$,

\Repeat {\ (for each step of episode)}

\State $a_t \leftarrow$ action $a_{i}$ with probability
$p(s_t,a_{i})$ for $s_t$

\State Take action $a_t$, observe reward $r_{t+1}$, and next state
$s_{t+1}$

\State \small{$Q(s_t,a_t)\leftarrow
Q(s_t,a_t)+\alpha_{t}\delta^Q_{t+1}$}

\State where \small{$\delta^Q_{t+1}=r_{t+1}+\gamma
max_{a'}Q(s_{t+1},a')-Q(s_t,a_t)$}

\State $p(s_t,a_{t})\leftarrow
p(s_t,a_{t})+k(r_{t+1}+max_{a'}Q(s_{t+1},a'))$

\State Normalize $\{p(s_{t},a_{i})|_{i=1,2,\ldots , m}\}$

\State $t\leftarrow t+1$

\Until {\ $s_{t+1}$ is terminal}

\Until {\ the learning process ends}

\end{algorithmic}
\end{algorithm}

Compared with basic Q-learning algorithms, the main feature of the PQL
algorithm is the straightforward probabilistic exploration strategy
and the reinforcement strategy is also applied to dynamically update
the probability distribution of action selection. The agent selects
actions based on the variable probability distribution over an
admissible action set at a certain state. Such an action selection
method keeps a proper chance of exploration instead of obeying the
policy learned so far, and makes a good tradeoff between exploration
and exploitation using probability.

\subsection{Fidelity-based PQL Algorithm}
PQL uses a probabilistic action selection method to improve the
exploration strategy and is an effective approach for stochastic
learning and optimization. But for most complex reinforcement
learning problems, the direction of achieving the objective is
always delayed due to the lack of feedback information during the
learning process unless the agent reaches the target state. Hence,
if we can extract more information from the system structure or
system behavior, the learning performance can be further improved
for complex learning problems with large learning space. Because
most of quantum control problems are complex and the concept of
fidelity is widely used in quantum information community
\cite{Nielsen and Chuang 2000}, \cite{Bason et al 2011},
\cite{Cozzini et al 2007}, we develop a fidelity-based PQL method
for learning control of quantum systems, which can also be applied
to some other complex RL problems.

The updating rule of fidelity-based PQL for $Q_{(s, a)}$ is the same
as (\ref{updaterule}). The probability distribution is updated for
each learning step. After the execution of action $a_{t}$ for state
$s=s_{t}$, the corresponding probability $p(s_{t},a_{t})$ is updated
according to the immediate reward $r_{t+1}$, the estimated value of
$Q_{(s_{t+1}, a')}$ for next state $s'=s_{t+1}$ and the fidelity
$F(s_{t+1}, s_{\text{target}})$ between the state $s_{t+1}$ and the
target state $s_{\text{target}}$. That is
\begin{equation}\small
p(s_{t},a_{t}) \leftarrow p(s_{t},a_{t})  +
k(r_{t+1}+\max_{a'}Q(s_{t+1},a')+F(s_{t+1}, s_{\text{target}})).
\end{equation}
The specification of the fidelity $F(s_{t+1}, s_{\text{target}})$
is defined regarding the objective of the learning control task.
In this study, a fidelity of quantum pure states (see Subsection
\ref{problemformulation}) is adopted for the learning control of
quantum systems. The parameter setting methods and the
normalization of the probability distribution of actions at state
$s=s_{t}$ are the same as that of PQL. The procedure of the
fidelity-based PQL algorithm is shown as \emph{Algorithm 2}.

\begin{algorithm}
\caption{Fidelity-based Probabilistic Q-learning} \label{FPQL}

\begin{algorithmic}[1]

\State Initialize $Q(s,a)$ arbitrarily

\State Initialize the policy \small{$\pi:
P^{\pi}=(p^{\pi}(s,a))_{n\times m}$} to be evaluated

\Repeat {\ (for each episode)}

\State Initialize $t=1$, $s_{t}$

\Repeat {\ (for each step of episode)}

\State $a_t \leftarrow$ action $a_{i}$ with probability
$p(s_t,a_{i})$ for $s_t$

\State Take action $a_t$, observe reward $r_{t+1}$, and next state
$s_{t+1}$

\State \small{$Q(s_t,a_t)\leftarrow
Q(s_t,a_t)+\alpha_{t}\delta^Q_{t+1}$}

\State where \small{$\delta^Q_{t+1}=r_{t+1}+\gamma
max_{a'}Q(s_{t+1},a')-Q(s_t,a_t)$}

\State \small{$p(s_t,a_{t})\leftarrow
p(s_t,a_{t})+k\delta^p_{t+1}$}

\State where
\footnotesize{$\delta^p_{t+1}=r_{t+1}+max_{a'}Q(s_{t+1},a')+F(s_{t+1},s_{\text{target}})$}

\State Normalize $\{p(s_t,a_{i})|_{i=1,2,\ldots , m}\}$

\State $t\leftarrow t+1$

\Until {\ $s_{t+1}$ is terminal}

\Until {\ the learning process ends}

\end{algorithmic}
\end{algorithm}

As for the convergence of FPQL, it is the same as that of basic
Q-learning \cite{Watkins and Dayan 1992}, because the difference
only lies in the exploration policy which does not affect the
convergence of the algorithms. Several constraints \cite{Watkins and
Dayan 1992}, \cite{Bertsekas and Tsitsiklis 1996} are listed in
Theorem 1 to ensure the convergence of FPQL.

\begin{theorem}[Convergence of FPQL]
Consider an FPQL agent in a nondeterministic Markov decision
process, for every state-action pair $s$ and $a$, the Q-value
$Q_{t}(s,a)$ will converge to the optimal state-action value
function $Q^*(s,a)$ if the following constraints are satisfied
\begin{enumerate}
\item The rewards in the whole learning process satisfy $(\forall
s,a)|r_{s}^{a}|\leq R$, where $R$ is a finite constant value;

\item A discount factor $\gamma \in [0,1)$ is adopted;

\item During the learning process, the nonnegative learning rate
$\alpha_t$ satisfies
\begin{equation}\label{equation 40}
\lim_{T\to \infty}\sum_{t=1}^{T}\alpha_t=\infty, \ \ \ \
\lim_{T\to \infty}\sum_{t=1}^{T}\alpha_{t}^2<\infty .
\end{equation}
\end{enumerate}
\end{theorem}

The difference between the proposed fidelity-based
probabilistic exploration strategy and the existing exploration
strategies can be explained from a point of view of physical mechanism.
As shown in Fig. 4, for a learning optimization problem, the
existing exploration strategies, e.g., $\epsilon$-greedy, softmax
and simulated annealing methods, apply a thermal fluctuation type of
exploration methods to acquire the chance of stepping over the local
optima; while the probabilistic exploration strategy is inspired by
quantum phenomena. It behaves like quantum tunnelling effect and can step
over the local optima in a more straightforward way. The direction
of the fidelity can strength this tunnelling effect. The physical
explanation can help demonstrate why the fidelity-based
probabilistic method may perform better for complex reinforcement
learning tasks.

\begin{figure}
\centering
\includegraphics[width=3.5in]{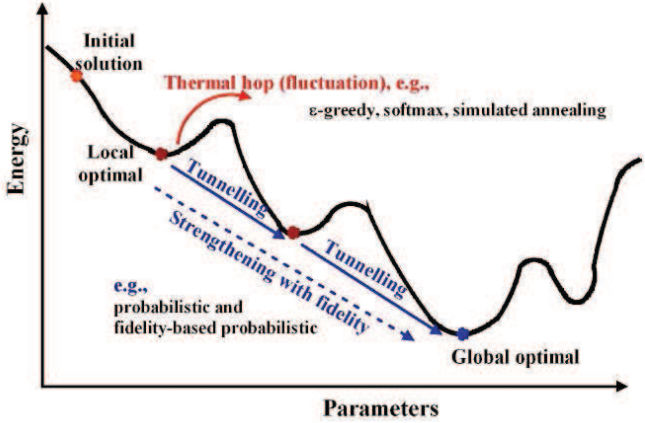}
\caption{A schematic of physical explanation and comparison of
different exploration strategies for optimization and learning
methods.}
\end{figure}

\begin{remark}
Although the fidelity-based PQL method is proposed for learning control
of quantum systems, it can also be applied to other RL problems when there is a clear and quantitative definition of
\emph{fidelity} that can effectively characterize the \emph{distance} between the
current state and the target state. The fidelity information can
speed up the learning process in that it shows the direction to the
target state and can help the learning process get out of the traps
when get lost. In addition, this fidelity signal is only used to
regulate the action selection probability distribution. It will not
deteriorate the exploration strategy of PQL.
\end{remark}

\begin{remark}
It is clear that the optimal policy of FPQL is closely related to
the probability distribution of the actions for each state, which
means the learning process essentially is also the process of
reducing the uncertainty of decisions on which action should be
chosen at a state. Hence, the characteristics of the FPQL algorithm
and its performance can also be described with the degree of
uncertainty for action selection. The measurement of uncertainty has
been well addressed through \emph{Shannon entropy} (i.e., Shannon
measure of uncertainty) \cite{Shannon and Weaver 1949} in
information science, where the amount of uncertainty is measured
by a probability distribution function $p$ on a finite set:
$S(p)=-\sum_{x\in X}p(x)\log_{2}p(x)$, where $X$ is the universal
set, $x$ is the element of the finite set $X$ and $p(x)$ is the
probability distribution function on $X$. Similar to Shannon
entropy, the concept of \emph{Exploration Entropy} can also be given
based on the probability distributions of actions to measure the
uncertainty of action selection. The resulting function is
$E_{(s)}=-\sum P_{(a_{i}|s)}\log_{2} P_{(a_{i}|s)}$. For an FPQL
system, the general uncertainty of action selection can be described
with the mean exploration entropy
$\overline{E_{(s)}}=\frac{\sum_{j=1,2,...,n}E_{(s_{j})}}{n}$, where
$S = \{s_{1}, s_{2}, ..., s_{n}\}$ is the state set. It is clear
that when all the probabilities are equal the exploration entropy
(uncertainty of action selection) will be maximum. In an FPQL
system, the maximum exploration entropy of a state $s$ with $m$
actions will be $\log_{2} m$. The maximum mean decision entropy
$\overline{E_{(s)}}$ should be obtained when all the action
probability distributions for each state are uniform, which is
always the situation at the initialization without any prior
knowledge about the environment. Along with the learning process,
$\overline{E_{(s)}}$ will tend to decrease and obtain its minimum
when the learning process converges and gives the optimal policy.
\end{remark}

\section{Fidelity-based PQL for learning control of quantum systems}

\subsection{Learning control of quantum
systems}\label{problemformulation} Learning control is an effective
method for quantum systems where a control law can be learned from experience and
the system performance can be optimized by searching for an optimal control
strategy in an iterative way \cite{Dong et al 2008b}, \cite{Rabitz et al 2000}-\cite{Granade et al 2012}. Here, we focus on the control
problem of quantum pure state transition for $N$-level quantum
systems \cite{Chen et al 2012}. Denote the eigenstates of the free
Hamiltonian $H_{0}$ of an $N$-level quantum system as
$D=\{|\phi_{i}\rangle \}_{i=1}^{N}$. An evolving state
$|\psi(t)\rangle$ of the controlled system can be expanded in terms
of the eigenstates in the set $D$:
\begin{equation}\label{superposition}
|\psi(t)\rangle=\sum_{i=1}^{N}c_{i}(t)|\phi_{i}\rangle
\end{equation}
where complex numbers $c_{i}(t)$ satisfy
$\sum_{i=1}^{N}|c_{i}(t)|^{2}=1$. We have the definition of
fidelity between two pure states.

\begin{definition}[Fidelity of Quantum Pure States]
The fidelity between two pure states
$|\psi^{a}\rangle=\sum_{i=1}^{N}c^{a}_{i}|\phi_{i}\rangle$ and
$|\psi^{b}\rangle=\sum_{i=1}^{N}c^{b}_{i}|\phi_{i}\rangle$ is
defined as
\begin{equation}\label{fidelity}
F(|\psi^{a}\rangle,
|\psi^{b}\rangle)=|\langle\psi^{a}|\psi^{b}\rangle|=|\sum^{N}_{i=1}(c^{a}_{i})^{*}c^{b}_{i}|,
\end{equation}
\end{definition}
where $(c^{a}_{i})^{*}$ is the complex conjugate of $c^{a}_{i}$.

Introducing a control $\varepsilon(t)\in L^2(\mathbf{R})$ acting on
the system via a time-independent interaction Hamiltonian $H_{I}$
and denoting $|\psi(t=0)\rangle$ as $|\psi_{0}\rangle$,
$C(t)=(c_i(t))_{i=1}^N$ evolves according to the Schr\"{o}dinger
equation~\cite{Dong and Petersen 2010IET}:
\begin{equation}\label{model}
\left \{
\begin{split}
& \iota\hbar\dot{C}(t)=[A+\varepsilon(t)B]C(t) \\
& C(t=0)=C_{0}
\end{split} \right.
\end{equation}
where $\iota=\sqrt{-1}$, $C_{0}=(c_{0i})_{i=1}^{N}$,
$c_{0i}=\langle\phi_{i}|\psi_{0}\rangle$,
$\sum_{i=1}^{N}|c_{0i}|^{2}=1$, $\hbar$ is the reduced Planck
constant, and the matrices $A$ and $B$ correspond to $H_{0}$ and
$H_{I}$, respectively. We assume that the $A$ matrix is diagonal and
the $B$ matrix is Hermitian \cite{Dong and Petersen 2010IET}. In
order to avoid trivial control problems we assume $[A, B]\equiv
AB-BA\neq 0$. Equation (\ref{model}) describes the evolution of a
finite dimensional control system. The propagator
$U(t_{1}\rightarrow t_{2})$ is a unitary operator such that for any
state $|\psi(t_{1})\rangle$ the state $|\psi(t_{2})\rangle =
U(t_{1}\rightarrow t_{2})|\psi(t_{1})\rangle$ is the solution at
time $t=t_{2}$ of~(\ref{superposition}) and~(\ref{model}) with the
initial condition $|\psi(t_{1})\rangle$ at time $t=t_{1}$.
$U(t_{1}\rightarrow t_{2})$ is also simplified as $U(t)$, $t\in
[t_{1},t_{2}]$, if the specific time $t_{1}$ can be neglected
when handling these problems. Assume that the control set
$\{\varepsilon_{j}, j=1,\dots, m\}$ is given. Every control
$\varepsilon_{j}$ corresponds to a unitary operator $U_{j}$. The
task of learning control is to find a control sequence
$\{\varepsilon_{l}, l=1, 2, 3, \dots \}$ where $\varepsilon_{l}\in
\{\varepsilon_{j}, j=1,\dots, m\}$ to drive the quantum system from
an initial state $|\psi_{0}\rangle$ to the target state
$|\psi_{f}\rangle$.

\begin{remark}
In the past decade, the research areas of quantum information
and machine learning have mutually benefitted from each other. On one hand,
quantum characteristics have been used for designing quantum or
quantum-inspired learning algorithms \cite{Dong et al 2008a},
\cite{Nielsen and Chuang 2000}, \cite{Malossini et al
2008}-\cite{Altman and Zapatrin 2010}. On the other hand, many traditional
learning algorithms have been applied for the control design of quantum
phenomena, including gradient-based algorithms \cite{Chakrabarti and
Rabitz 2007}, \cite{Roslund and Rabitz 2009}, genetic algorithm (GA)
\cite{Rabitz et al 2000}, \cite{Tsubouchi and Momose 2008} and fuzzy logic
\cite{Chen et al 2012}. For example, gradient-based methods have
been widely used in model-based control design and theoretical
analysis of quantum systems. Since we assume that very limited control resources are available,
gradient-based algorithms cannot be applied to the quantum control problem in this paper.
GA methods have achieved great success
for quantum learning control in laboratory \cite{Rabitz et al 2000}. However, a large amount of experimental data
is required to optimize the control performance since the closed-loop learning process involves the collection of experimental data
and the searching of optimized pulses based on the updating of experimental data \cite{Rabitz et al 2000}. In this paper, we consider a class of quantum control problems with a limited set of control fields. This class of problems is significant in quantum control since different constraints are common for quantum control systems. We formulate this class of quantum control
problems as a model-free sequential Markovian
decision process (MDP). Reinforcement learning is a good candidate to solve a MDP
problem. Hence, we apply the proposed FPQL
approach to this class of quantum control problems that can be used to test
the effectiveness of FPQL as well as to provide an effective design
approach for quantum systems with limited control resources.
\end{remark}

\subsection{Quantum controlled transition landscapes}
Learning control of quantum systems aims to find an optimal control
strategy to manipulate the dynamics of physical processes on the
atomic and molecular scales \cite{Rabitz et al 2000}. In recent
years, quantum control landscapes \cite{Rabitz et al 2004},
\cite{Chakrabarti and Rabitz 2007} provide a theoretical footing for
analyzing learning control of quantum systems. A control landscape
is defined as the map between the time-dependent control Hamiltonian
and associated values of the control performance functional. Most
quantum control problems can be formulated as the maximization of an
objective performance function. For example, as shown in Fig. 5, the
performance function $J(\varepsilon)$ is defined as the functional
of the control strategy $\varepsilon={\varepsilon_{i},
i=1,2,...,M}$, where $M$ is a positive integer that indicates the
number of the control variables ($M=2$ for the case shown in Fig.
5).

\begin{figure}
\centering
\includegraphics[width=3.8in]{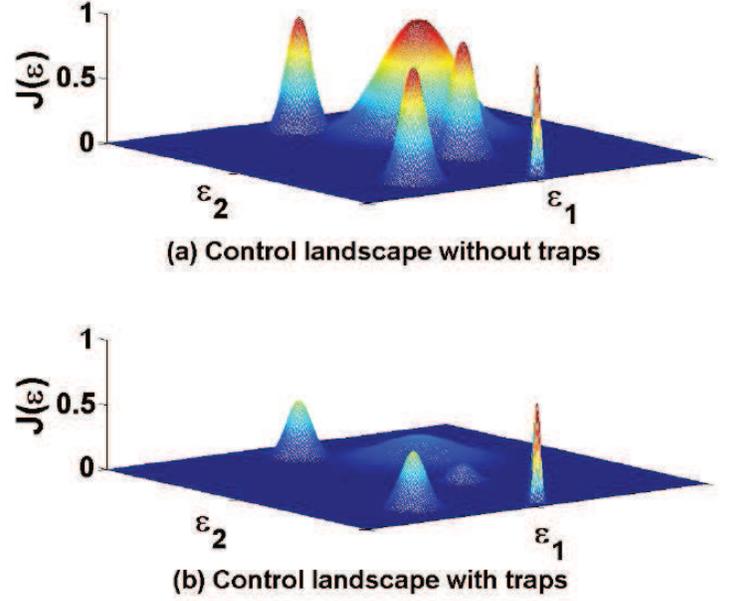}
\caption{An example of quantum control landscapes. (a) Quantum
control landscape without traps (local maxima) where all the peaks
are of the same height and thus all of them are global maxima; (b)
Quantum control landscape with traps where the
landscape has one highest peak representing the global maximum and
several peaks of lower height corresponding to local maxima.}
\end{figure}

Although quantum control applications may span a variety of
objectives, most of them correspond to maximizing the probability of
transition from an initial state $|\psi_{0}\rangle$ to a desired
final state $|\psi_{f}\rangle$ \cite{Rabitz et al 2004}. For the
state transition problem with $t\in [0, T]$, we define the quantum controlled
transition landscape as

\begin{equation}\label{dynamiclandscape}
J(\varepsilon)=\text{tr}(U_{(\varepsilon,
T)}|\psi_{0}\rangle\langle\psi_{0}|U^{\dag}_{(\varepsilon,
T)}|\psi_{f}\rangle\langle\psi_{f}|),
\end{equation}
where $\text{tr}(\cdot)$ is the trace operator and $U^{\dag}$ is the
adjoint of $U$. The objective of the learning control system is to
find a global optimal control strategy $\varepsilon^*$ which
satisfies
\begin{equation}
\varepsilon^*=\text{argmax}_{\varepsilon}J(\varepsilon).
\end{equation}

If the dependence of $U_{(T)}$ on $\varepsilon$ is suppressed (see \cite{Chakrabarti and Rabitz
2007}),
(\ref{dynamiclandscape}) can be reformulated as
\begin{equation}\label{kinematiclandscape}
J(U)=\text{tr}(U_{(T)}|\psi_{0}\rangle\langle\psi_{0}|U^{\dag}_{(T)}|\psi_{f}\rangle\langle\psi_{f}|).
\end{equation}
Equations (\ref{dynamiclandscape}) and (\ref{kinematiclandscape})
are called the dynamic control landscape (denoted as
$J_D(\varepsilon)$ instead) and the kinematic landscape (denoted as
$J_K(U)$ instead), respectively (see \cite{Chakrabarti and Rabitz
2007}).

The characteristic of the existence or absence of traps is most
important for exploring the quantum control landscape with a
learning control algorithm, which can be studied using critical
points. A dynamic critical point is defined by
\begin{equation}
\nabla J_D(\varepsilon)=\delta J_D(\varepsilon)/\delta \varepsilon=0
\end{equation}
and a kinematic critical point is defined by
\begin{equation}
\nabla J_K(U)=\delta J_K(U)/\delta U=0
\end{equation}
where $\nabla$ denotes gradient. By the chain rule, we have
\begin{equation}
\nabla J_D(\varepsilon)=\frac{\delta J_K(U)}{\delta U_{(\varepsilon,
T)}}\frac{\delta U_{(\varepsilon, T)}}{\delta \varepsilon}=\nabla
J_K(U)\frac{\delta U_{(\varepsilon, T)}}{\delta \varepsilon}.
\end{equation}

According to the results in \cite{Chakrabarti and Rabitz 2007},
we can summarize the properties of quantum controlled transition
landscape as Theorem 2.

\begin{theorem}\label{Theorem1}
For the quantum control problem defined with the dynamic control
landscape (\ref{dynamiclandscape}) and the kinematic control
landscape (\ref{kinematiclandscape}), respectively, the properties
of the solution sets of the quantum controlled transition landscape
are listed as follows:
\begin{enumerate}
\item The kinematic control landscape is free of traps (i.e., all
critical points of $J_K(U)$ are either global maxima or saddles) if
the operator $U$ can be any unitary operator (i.e., the system is
completely controllable);

\item The dynamic control landscape is free of traps if
(\romannumeral1) the operator $U$ can be any unitary operator and
(\romannumeral2) the Jacobian $\delta U_{(\varepsilon, T)}/\delta
\varepsilon$ has full rank at any $\varepsilon$.
\end{enumerate}
\end{theorem}

For detailed proof and discussion about Theorem
\ref{Theorem1}, please refer to \cite{Rabitz et al 2004},
\cite{Chakrabarti and Rabitz 2007}.

\begin{remark}
The quantum controlled transition landscape theory is the
theoretical foundation for learning control design. The FPQL algorithm
has potential for quantum learning control problems. The reasons can
be stated from three aspects: (\romannumeral1) The probabilistic
action selection method makes a better balance between exploration
and exploitation, since too much exploitation is easy to be trapped
and too much exploration will deteriorate the learning performance;
(\romannumeral2) The theoretical analysis of the solution sets of a
quantum control landscape can help design a fidelity-based method to
improve the learning performance; (\romannumeral3) The learning
scheme of reinforcement learning is more suitable for model-free
real laboratory applications than typical gradient-based methods
which need specific models.
\end{remark}

In the next two subsections, the learning control problems of a
spin-$\frac{1}{2}$ system and a $\Lambda$-type atomic system are
studied using the proposed FPQL algorithm, which shows that FPQL is
an alternative effective approach for quantum control design.

\subsection{Example 1: learning control of a spin-$\frac{1}{2}$
quantum system}
The
spin-$\frac{1}{2}$ system is a typical 2-level quantum system and
has important theoretical implications and practical applications.
Its Bloch vector can be visualized on a 3D Bloch sphere as shown in
Fig. 6. The state of the spin-$\frac{1}{2}$ quantum system
$|\psi\rangle$ can be represented as
\begin{equation}
|\psi\rangle=\cos\frac{\theta}{2}|0\rangle+e^{i\varphi}\sin\frac{\theta}{2}|1\rangle
\end{equation}
where $\theta\in[0,\pi]$ and $\varphi\in[0,2\pi]$ are polar angle
and azimuthal angle, respectively, which specify a point
$\overrightarrow{a}=(x,y,z)=(\sin\theta\cos\varphi,\sin\theta\sin\varphi,\cos\theta)$
on the unit sphere in $\mathbf{R}^{3}$.

\begin{figure}
\centering
\includegraphics[width=3.8in,height=4.2in]{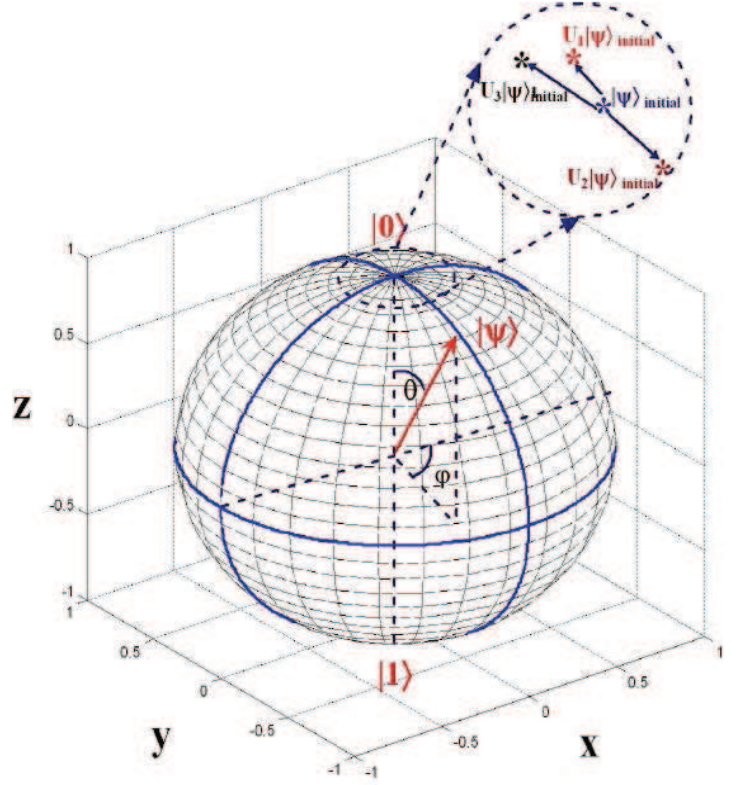}
\caption{Demonstration of a spin-$\frac{1}{2}$ system with a Bloch
sphere in a 3D Cartesian  coordinates and the state transitions for
an initial quantum state $|\psi\rangle_{initial}$ using different one-step
controls ($U_{1}$,$U_{2}$,$U_{3}$)} \label{presult}
\end{figure}

At each control step, the permitted controls for every state are
$U_{1}$ (no control input), $U_{2}$ (a positive pulse control) and
$U_{3}$ (a negative pulse control). Fig. 6 shows a sketch map of
one-step control effects on the evolution of the quantum system.
The propagators $\{U_{i},i=1,2,3\}$ are listed as follow:
\begin{equation}
U_{1}=e^{-iI_{z}\frac{\pi}{15}} \label{Ix},
\end{equation}
\begin{equation}
U_{2}=e^{-i(I_{z}+0.5I_{x})\frac{\pi}{15}} \label{Ix},
\end{equation}
\begin{equation}
U_{3}=e^{-i(I_{z}-0.5I_{x})\frac{\pi}{15}} \label{Ix},
\end{equation}
where
\begin{equation}
I_{z}=\frac{1}{2}
\begin{pmatrix}
  1    & 0   \\
  0   & -1   \\
  \end{pmatrix}, \ \
  I_{x}=\frac{1}{2}
\begin{pmatrix}
  0    & 1   \\
  1   & 0   \\
  \end{pmatrix} \label{Ix}.
\end{equation}

\begin{figure*}
\centering
\includegraphics[width=6.5in]{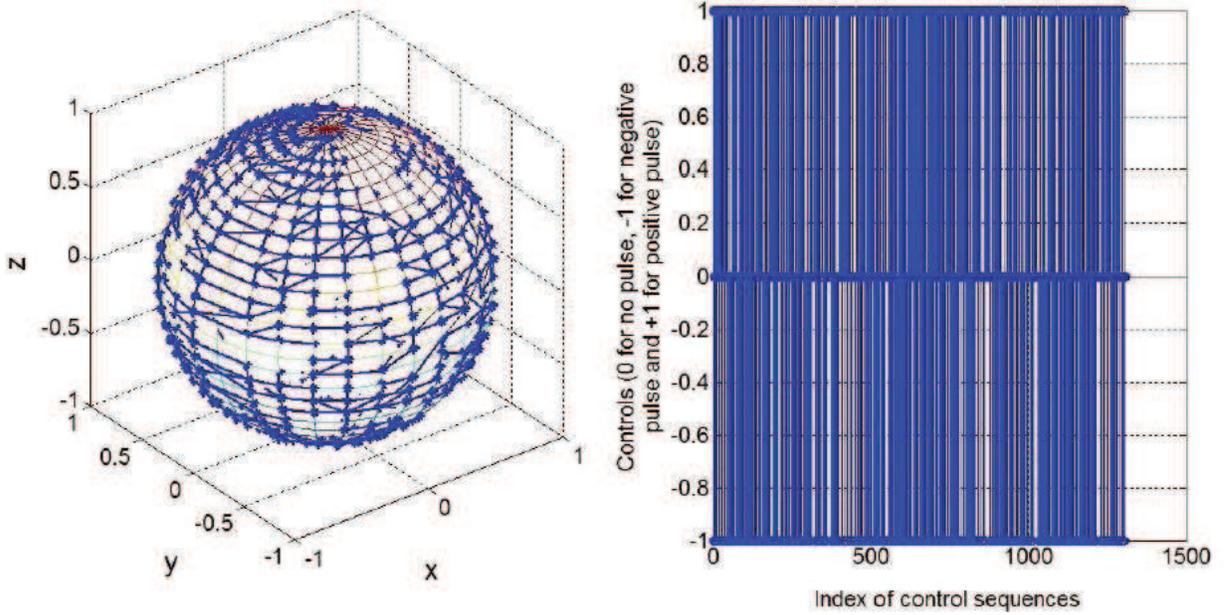}
\caption{Demonstration of a stochastic control case without
learning. The left figure shows the state transition path and the
right figure shows the control sequence used (0 for no
pulse, -1 for negative pulse and +1 for positive pulse)}
\label{presult}
\end{figure*}

Now the control objective is to control the spin-$\frac{1}{2}$
system from the initial state
$(\theta=\frac{\pi}{60},\varphi=\frac{\pi}{30})$ to the target state
$(\theta=\frac{41\pi}{60},\varphi=\frac{29\pi}{30})$ with minimized
control steps. Fig. 7 shows one of the control process before
learning, where the controls are selected randomly and after a long
control sequence the system state may be transited to the target
state. We apply the fidelity-based PQL, PQL and QL algorithms to
this learning control problem, respectively. Now we reformulate the
RL problem of controlling a quantum system from an initial state
$s_{\text{initial}}=|\psi_{\text{initial}}\rangle$ to a desired
target state $s_{\text{target}}=|\psi_{\text{target}}\rangle$ as
follows: the state set is
$S=\{s_{i}=|\psi_{\text{i}}\rangle\},i=1,2,\ldots,n$ and the action
set is $A=\{a_{j}=u_{j}\},j=1,2,\ldots,m$. The experiment settings
for these algorithms are listed as follows: $r=-1$ for each control
step until it reaches the target state, then it gets a reward of
$r=1000$; the discount factor $\gamma=0.99$, the learning rate
$\alpha=0.01$ and the Q-values are all initialized as 0. For PQL and
fidelity-based PQL, $k=0.01$. The $\epsilon$-greedy exploration
strategy is used and $\epsilon=0.1$.

\begin{figure*}
\centering
\includegraphics[width=7.3in,height=2.3in]{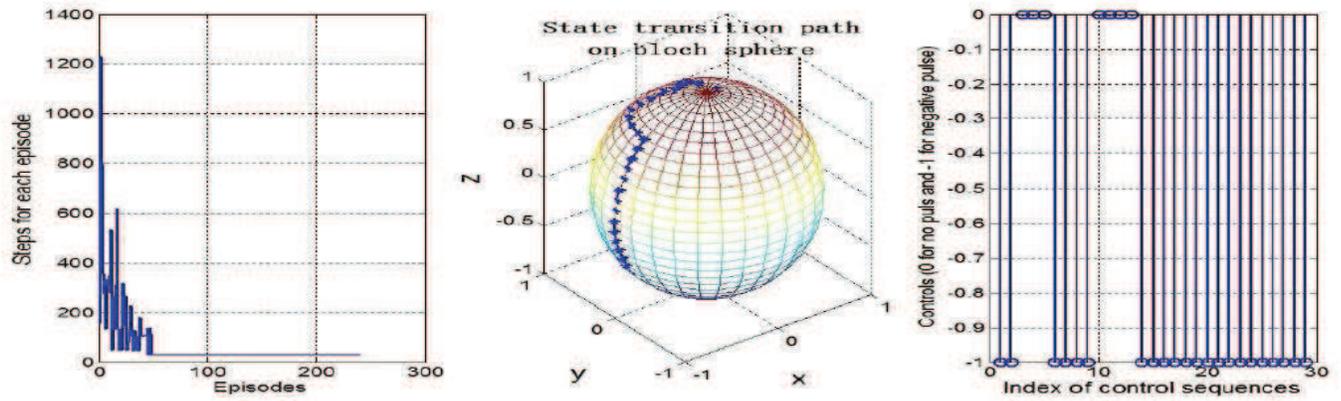}
\caption{Learning performance of fidelity-based PQL and the
learning results with an optimal control sequence} \label{presult}
\end{figure*}

\begin{figure*}
\centering
\includegraphics[width=7.3in,height=2.3in]{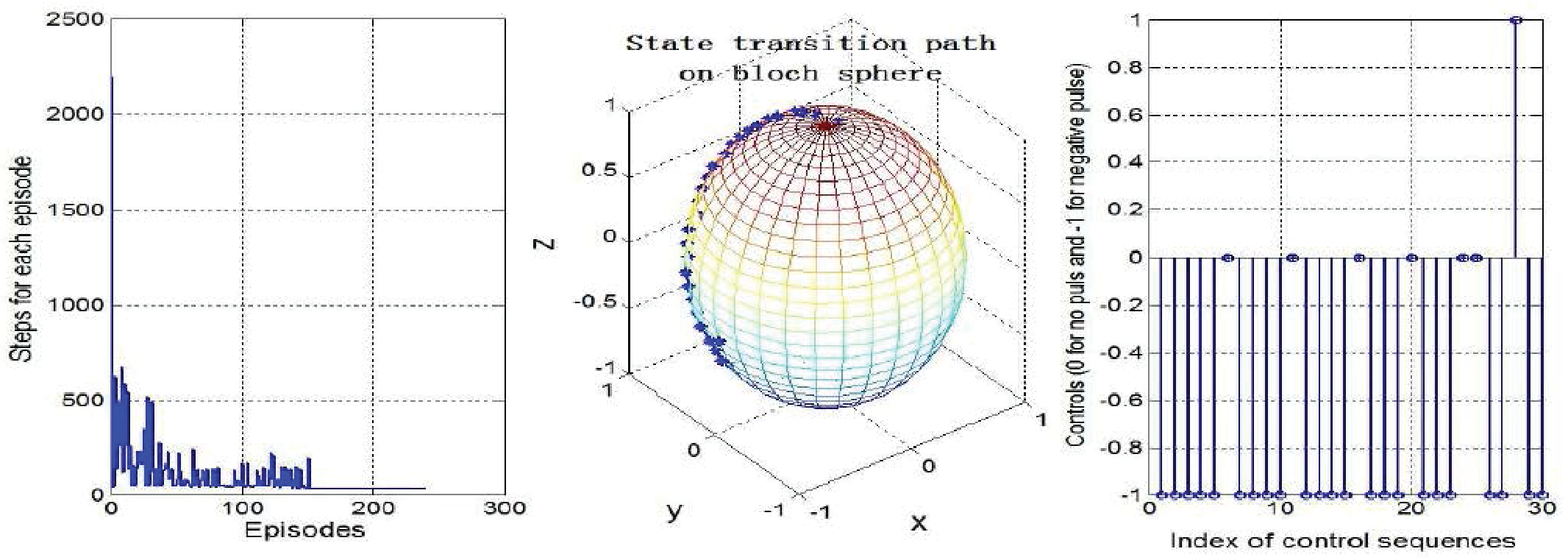}
\caption{Learning performance of PQL and the learning results with
an optimal control sequence} \label{presult}
\end{figure*}

\begin{figure*}
\centering
\includegraphics[width=7.3in,height=2.3in]{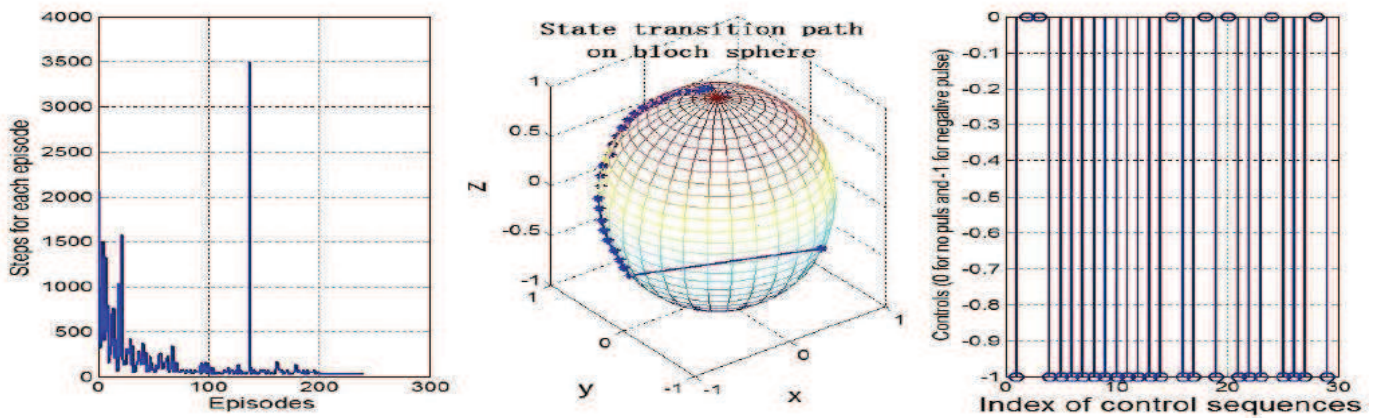}
\caption{Learning performance of standard QL with
$\varepsilon$-greedy policy and the learning results with an
optimal control sequence} \label{presult}
\end{figure*}

Figs. 8-10 show the control performance of all these algorithms,
respectively. Hundreds of times of learning process are carried out
for each experiment and all the results maintain similar
performance. We provide the results of one time of learning process for each experiment. The experimental results show that fidelity-based PQL
outperforms PQL and standard QL. The fidelity-based PQL quickly find
the optimal control sequence after less than 50 episodes, while PQL
needs about 150 episodes and QL needs more than 200 episodes. For
each episode in the learning process, QL and PQL also need much more
steps to find the target state. A clearer performance comparison
between the fidelity-based PQL, PQL and QL is shown as in Fig. 11.
It is clear that the fidelity-based method contributes to more
effective tradeoff between exploration and exploitation than PQL and
confines it from exploring too much in an economical way with
respect to exploration cost. Although the fidelity-based PQL needs a
little more steps in the early
learning stage
(which makes its performance lies between QL and PQL), it can quickly converge to the optimal policy and
remarkably outperforms both of QL and PQL. The final control results
with the learned optimal control sequence that
controls the spin-$\frac{1}{2}$ quantum state from the initial state
to the target state is demonstrated in Fig. 12.

\begin{figure}
\centering
\includegraphics[width=3.5in]{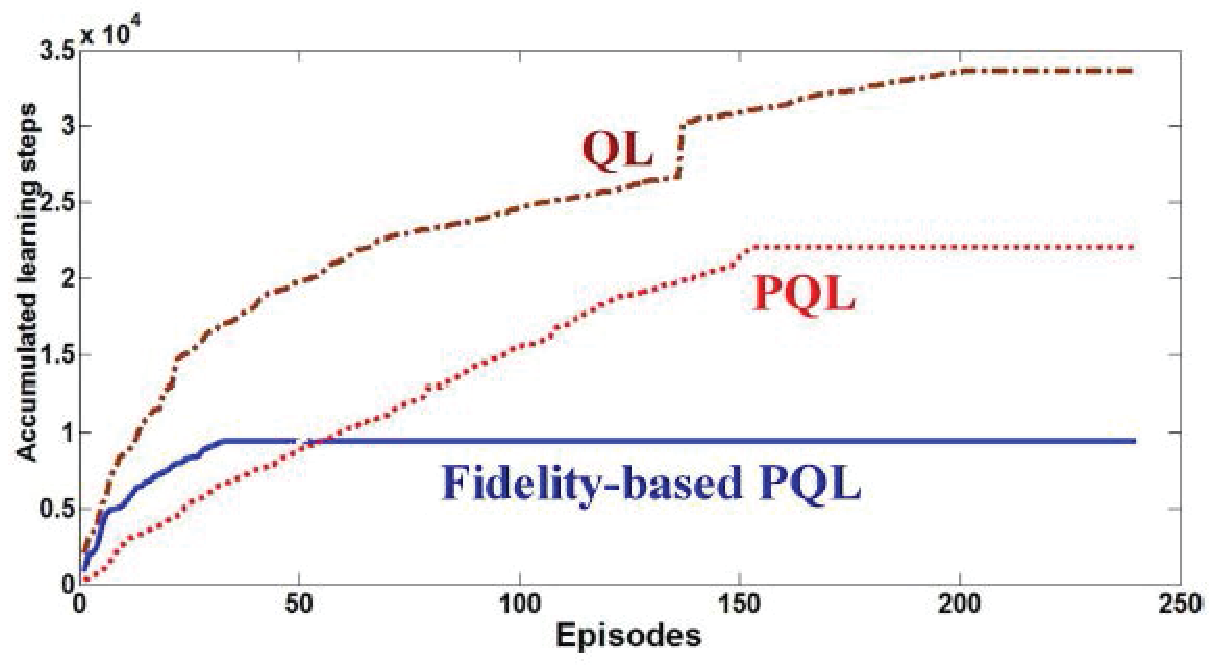}
\caption{Comparison of Learning performances between Fidelity-based
PQL, PQL and QL} \label{presult}
\end{figure}

\begin{figure}
\centering
\includegraphics[width=3.2in,height=3.2in]{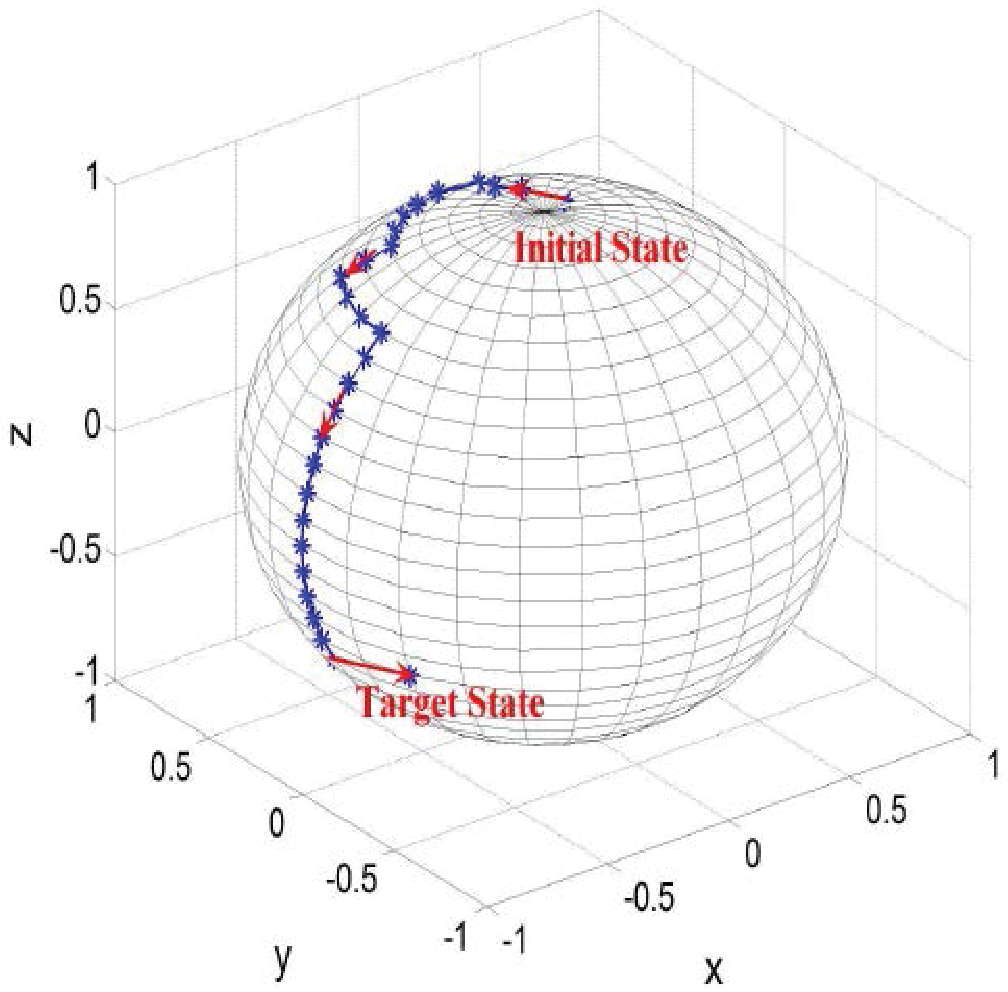}
\caption{The control results with the learned optimal control
sequence} \label{presult}
\end{figure}

\subsection{Example 2: learning control of a $\Lambda$-type
quantum system}

Now we consider a $\Lambda$-type atomic system and demonstrate the
fidelity-based PQL design process. The three level $\Lambda$-type
atomic system is a representative of the multi-level system, which
has wide applications in the fields of chemistry, quantum
physics and quantum information \cite{You and Nori 2011}. For the $\Lambda$-type system
shown in Fig. 13, the evolving state $|\psi(t)\rangle$ can be
expanded in terms of the eigenstates as follows:

\begin{figure}
\centering
\includegraphics[width=1.9in]{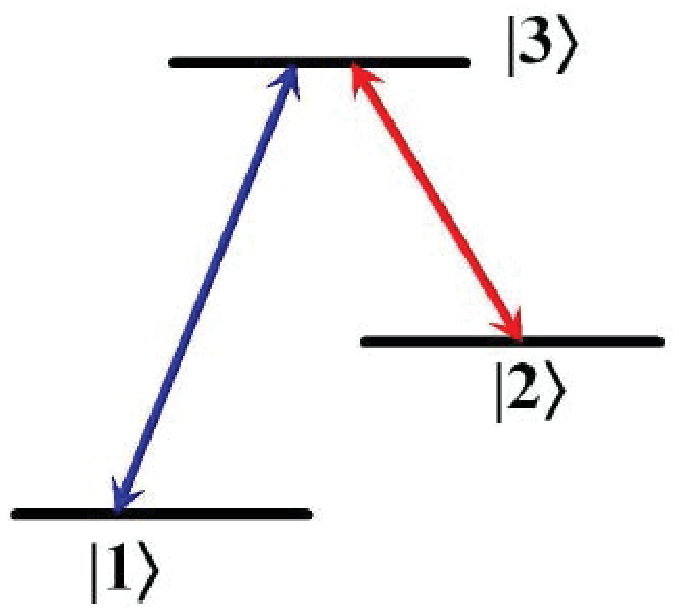}
\caption{A schematic of a 3-level $\Lambda$-type atomic system.}
\label{level3}
\end{figure}

\begin{equation}\label{level3system1}
|\psi(t)\rangle=
c_{1}(t)|1\rangle+c_{2}(t)|2\rangle+c_{3}(t)|3\rangle,
\end{equation}
where $|1\rangle$, $|2\rangle$ and $|3\rangle$ are the basis states
of the lower, middle and upper atomic states, respectively. At each
control step, the permitted controls are a finite number of
(positive or negative) control pulses, i.e., we have the propagators
\begin{equation}\label{level3system2}
U_E=e^{-i\Delta t(H_0+0.1EH_1)}
\end{equation}
where $\Delta t=0.1$,
\begin{equation}\small \label{level3system3}
H_{0}=
\left(%
\begin{array}{ccc}
  1.5 & 0 & 0 \\
  0 & 1  & 0 \\
  0 & 0  & 0 \\
\end{array}%
\right), \ H_{1}=
\left(%
\begin{array}{ccc}
  0 & 0 & 1 \\
  0 & 0  & 1 \\
  1 & 1  & 0 \\
\end{array}%
\right),
\end{equation}
and $E\in\{0,\pm1,\pm2,\ldots,\pm20\}$ is the number of
chosen control pulses at a certain control step.

Now the control objective is to control the $\Lambda$-type atomic
system from the initial state
$|\psi_{\text{initial}}\rangle=(1,0,0)$ to the target state
$|\psi_{\text{target}}\rangle=(0,0,1)$ with a fixed number of
control steps. We apply the fidelity-based PQL, PQL and QL
algorithms to this learning control problem, respectively. First
we reformulate the RL problem of controlling a quantum system from
an initial state
$s_{\text{initial}}=|\psi_{\text{initial}}\rangle$ to a desired
target state $s_{\text{target}}=|\psi_{\text{target}}\rangle$ as
follows: the number of control steps is fixed as a constant number
of $100$, so that we can use a virtual state set to construct the
state-action space instead of the real state space (with a very
high dimension) of the $\Lambda$-type system and the state set
$S=\{s_{i}\},i=1,2,\ldots,101$ and the action set is
$A=\{a_{j}=E_j=j-21\}, j=1,2,\ldots,41$. The experiment settings
for these algorithms are listed as follows: $r=0$ for each control
step until it reaches the target state at the end of the control
process where it gets a reward of $r=1000$; the discount factor
$\gamma=0.99$, the learning rate $\alpha=0.01$ and the Q-values
are all initialized as 0. For PQL and fidelity-based PQL,
$k=0.01$. The $\epsilon$-greedy exploration strategy is used for
QL and $\epsilon=0.1$. The fidelity for a current policy $\pi$ is
defined as
$F=|\langle\psi^{\pi}_{f}|\psi_{\text{target}}\rangle|$.

\begin{figure}
\centering
\includegraphics[width=3.2in]{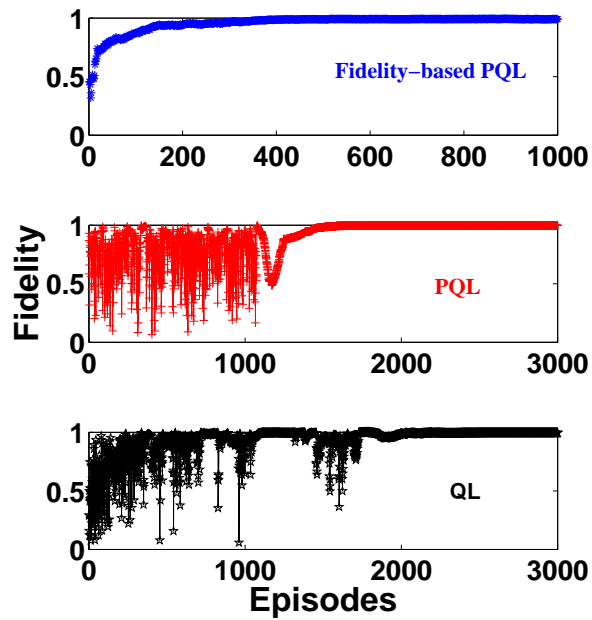}
\caption{Control performances with respect to the fidelity between
the final state and the target state using fidelity-based PQL, PQL
and QL, respectively.} \label{level3compare}
\end{figure}

The learning performances of fidelity-based PQL, PQL and QL are
shown in Fig. 14 with respect to fidelity, which is one of all
the alike results for hundreds of experiments we carried out. The
learning process converges after about $300$ episodes using
fidelity-based PQL, while PQL needs about $1450$ episodes and QL
needs about $2000$ episodes. The oscillation for PQL and QL before
the learning processes converge in Fig. 14 is due to the performance
criteria regarding fidelity instead of accumulated learning steps as
used in Fig. 11. The performance shown in Fig. 14 is very sensitive
to the exploration behavior in the state-action space for PQL and
QL, while the fidelity-based method shows an almost monotonically
improved learning behavior.

The final optimal control sequence is shown in Fig.
15. With this learned optimal control sequence the $\Lambda$-type
atomic system described by Equations
(\ref{level3system1})-(\ref{level3system3}) is controlled from the
initial state $|\psi_{\text{initial}}\rangle=(1,0,0)$ to the target
state $|\psi_{\text{target}}\rangle=(0,0,1)$ and the population
evolution trajectories are demonstrated in Fig. 16. All these
numerical results demonstrate the success of the proposed
fidelity-based PQL method. In addition, the learning control of the
$\Lambda$-type atomic system can also be implemented with policy
iteration \cite{Xu et al 2007}, \cite{Xu et al 2011}, but it is out
of the scope of this paper. More comparison results between the
value iteration and policy iteration for reinforcement learning will
be presented in our future work.

\begin{figure}
\centering
\includegraphics[width=3.2in]{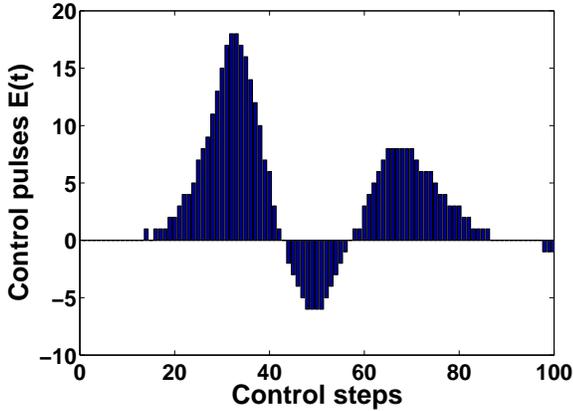}
\caption{The learned optimal control pulse sequence.} \label{level3control}
\end{figure}

\begin{figure}
\centering
\includegraphics[width=3.2in]{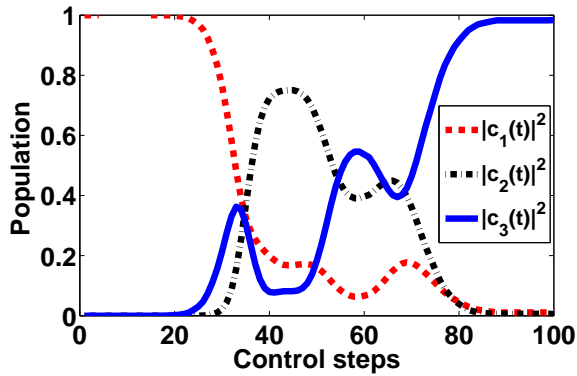}
\caption{The population evolution trajectories with the learned
optimal control pulse sequence.} \label{level3track}
\end{figure}

\section{Conclusions}
In this paper, a probabilistic action selection method is introduced
for Q-learning and an FPQL algorithm is presented for the learning
control design of quantum systems. In FPQL, the fidelity information
can be extracted from the system structure or the system behavior.
The aim is to design a good exploration strategy for a better
tradeoff between exploration and exploitation and to speed up the
learning process as well. The experimental results show that FPQL is
superior to basic Q-learning with respect to convergence speed. The
control problems of a spin-$\frac{1}{2}$ system and a $\Lambda$-type
atomic system are adopted to demonstrate the performance of FPQL.
Although all the cases we considered in this study are discrete
examples, which are most widely used in practical applications, the
proposed fidelity-based probabilistic action selection method can be
extended to other reinforcement learning algorithms and applications
using function approximation with a continuous probability
distribution and related iteration methods. In addition, our future
work will focus on further comparison of FPQL with other
existing learning methods (e.g., GA, gradient-base methods
and neural networks) for more general quantum control problems.

\end{document}